\documentclass{article}
\usepackage{spconf,amsmath,graphicx,bm}

\usepackage{amssymb}
\usepackage{booktabs}
\usepackage[pagebackref,breaklinks,colorlinks]{hyperref}
\usepackage[sort&compress,numbers,comma]{natbib}
\usepackage{balance}
\usepackage{setspace}

\usepackage[normalem]{ulem}
\usepackage{xcolor}

\def \eg {\emph{e.g.}}

\title{Video Question Answering using Clip-guided Visual-text Attention}

\name{Shuhong Ye$^{\star}$$^\$$, Weikai Kong$^{\star}$$^{\$}$,  Chenglin Yao$^{\star}$, Jianfeng Ren$^{\star\dagger}$, Xudong Jiang $^{\ddagger}$ 
\thanks{$^{\$}$ The authors contributed equally.}
\thanks{This work was supported in part by the National Natural Science Foundation of China under Grant 72071116, and in part by the Ningbo Municipal Bureau Science and Technology under Grants 2019B10026 and 2022Z173.}}
\address{${}^{\star}$School of Computer Science, University of Nottingham Ningbo China\\
$^{\dagger}$Nottingham Ningbo China Beacons of Excellence Research and Innovation Institute, \\
University of Nottingham Ningbo China, China\\
$^{\ddagger}$School of Electrical \& Electronic Engineering, Nanyang Technological University}

\begin{document}
%
\maketitle
\begin{abstract}
Cross-modal learning of video and text plays a key role in Video Question Answering (VideoQA). 
In this paper, we propose a visual-text attention mechanism to utilize the Contrastive Language-Image Pre-training (CLIP) trained on lots of general domain language-image pairs to guide the cross-modal learning for VideoQA. Specifically, we first extract video features using a TimeSformer and text features using a BERT from the target application domain, and utilize CLIP to extract a pair of visual-text features from the general-knowledge domain through the domain-specific learning. 
We then propose a Cross-domain Learning to extract the attention information between visual and linguistic features across the target domain and general domain. The set of CLIP-guided visual-text features are integrated to predict the answer. 
The proposed method is evaluated on MSVD-QA and MSRVTT-QA datasets, and outperforms state-of-the-art methods.

\end{abstract}
\section{Introduction}
\label{sec:intro}



Video Question Answering (VideoQA) has become increasingly popular in 
vision-language navigation \cite{anderson2018vision_navi}, multimedia recommendation \cite{chen2017attentive_application_multimedia_recommendation}, and communication systems~\cite{bica2016mutual_icassp}. 
The target is to correctly answer questions about a video, which requires a deep understanding of video scenes, question semantics, and fine-grained vision-language alignment. 
Many methods have been developed for cross-modality learning~\cite{lei2018tvqa, lei2020tvqa+,  lei2021less_VLP_Clipbert_benchmark, guzhov2022audioclip_icassp_multi_modality}. 
But due to limited training data, some linguistic concepts in the answer space have no corresponding video samples, resulting in the lack of linguistic supervision. Such a problem hinders accurate pairing between linguistic features and corresponding visual features, and hence limits the cross-modality learning ability of the existing models \cite{xu2017video_AMU_attention_model_benchmark,gao2018motion_CoMem_mem_memory_mode_benchmark,fan2019heterogeneous_HME_mem_att_memory_model_benchmark,jiang2020reasoning_HGA_graph_model_benchmark,jiang2020divide_QUEST_att_attention_model_benchmark, le2022hierarchical_HCRN_relation_model_benchmark, xiao2022video_HQGA_graph_model_benchmark}. 

There are two main directions for improving VideoQA models. 
1) Exploit deeper correlations from annotated video-text pairs using recurrent neural network \cite{gao2018motion_CoMem_mem_memory_mode_benchmark,fan2019heterogeneous_HME_mem_att_memory_model_benchmark}, graph neural network \cite{xiao2022video_HQGA_graph_model_benchmark,jiang2020reasoning_HGA_graph_model_benchmark, wang2021dualvgr_graph_model_benchmark}, conditional relation network \cite{le2022hierarchical_HCRN_relation_model_benchmark}, or attention-based models \cite{xu2017video_AMU_attention_model_benchmark, jiang2020divide_QUEST_att_attention_model_benchmark}. 
2) Overcome the challenge of insufficient linguistic supervision by importing general domain knowledge from large-scale pre-trained vision-language models, \eg, HERO \cite{li2020hero_VLP}, ClipBert \cite{lei2021less_VLP_Clipbert_benchmark}, and JustAsk \cite{yang2021just_VLP_JustAsk_benchmark}. 
Benefiting from the additional general domain knowledge that can describe unseen answers in the target application domain, pre-trained models significantly boost the performance of downstream VideoQA tasks and achieve state-of-the-art results on many VideoQA datasets \cite{xu2017video_AMU_attention_model_benchmark, yu2018joint_JFusion_MSRVTT-MC, lei2018tvqa}. 

It should be noted there may be a knowledge discrepancy between the target domain and the general domain. Failing to bridge the discrepancy may bring conflicts in the cross-modal knowledge representation. To address this problem, we propose a two-stage cross-domain cross-modal learning framework under the guidance of CLIP model \cite{radford2021learning_CLIP}. CLIP is pre-trained from large-scale image-text pairs so that more unseen answers can be described. It provides a bidirectional transform of features from the salient contents in video frames and text captions. In the first stage, vision-language features are extracted in a domain-specific way. In the target domain, the video features are extracted via a TimeSformer \cite{bertasius2021space_timesformer} and the question and answer texts are encoded using transformers \cite{vaswani2017attention_transformer_TF,VIT}. To incorporate the general domain knowledge from the large-scale pre-trained CLIP, the key video frames are extracted as the salient contents and fed into the CLIP to generate the CLIP-guided visual features, and the question and answers are fed into the CLIP to generate the CLIP-guided linguistic features. In the second stage of cross-domain learning, to bridge the knowledge discrepancy, four CLIP-guided visual-text encoders are designed to exploit the cross-modal cross-domain attention information. Finally, the four sets of visual-text features are fused to predict the answer.

Our contributions are three-fold: 1) The proposed method effectively extracts the visual and linguistic features from the general domain knowledge using the pre-trained CLIP and from the target domain using TimeSformer and transformer. 
2) The proposed CLIP-guided visual-text attention mechanism effectively integrates the general domain knowledge into the target domain for cross-modal cross-domain learning in VideoQA. 
3) Experimental results on two large benchmark datasets demonstrate that the proposed method significantly outperforms state-of-the-art VideoQA models. 

\begin{figure*}[htpb]
\centering
\includegraphics[width=0.96\textwidth]{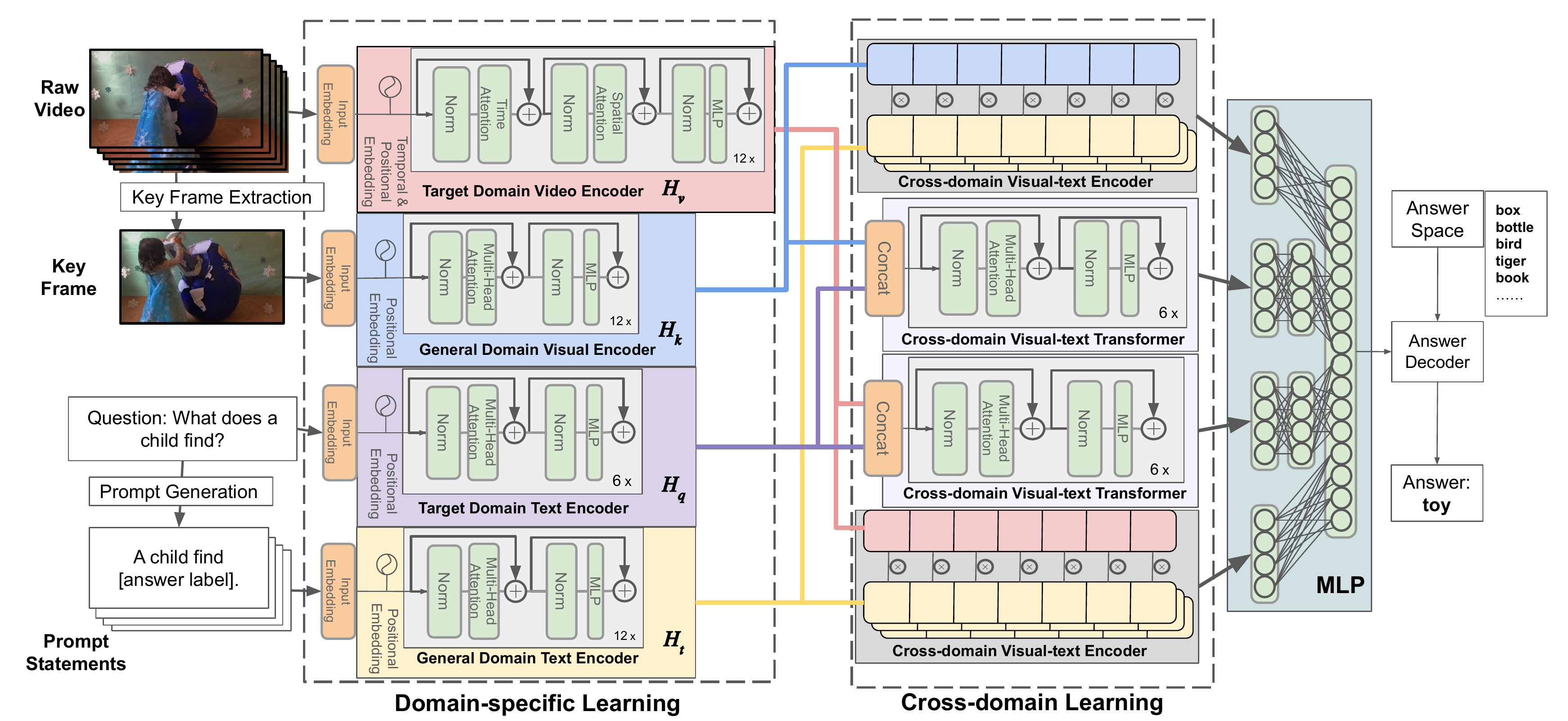}
  \caption{
  Overview of the proposed CCVQA. It consists of two main modules. 1) Domain-specific Learning, including a video encoder to extract spatial-temporal features $\bm{H}_{v}$, a text encoder to encode question descriptions $\bm{H}_{q}$ from the target domain, a CLIP-guided video key frame encoder to encode object representations $\bm{H}_{k}$ with language supervision, and a CLIP-guided candidate answer encoder to generate linguistic features $\bm{H}_{t}$ with vision supervision in the general domain. 2) Cross-domain Learning, including four sets of CLIP-guided visual-text encoders to model cross-domain cross-modal feature interaction through the attention mechanism. Finally, the features are fused using an MLP and fed to a decoder to predict the answer.}
  \label{fig:2}
\end{figure*}

\section{Proposed Method}

\subsection{Overview of Proposed Method}
The block diagram of the proposed CLIP-guided Cross-domain Video Question Answering (CCVQA) model is shown in Fig~\ref{fig:2}. It consists of two main modules.  
1) Domain-specific Learning. The vision and language features are first extracted in a domain-specific way. More specifically, a TimeSformer \cite{bertasius2021space_timesformer} is utilized to extract the visual features from the video sequences in the target VideoQA domain. To incorporate the general domain knowledge, the key frames of the video are first selected and fed into the image stream of CLIP \cite{radford2021learning_CLIP} to extract visual features that are compatible with language supervision. 
Similarly, the transformer \cite{vaswani2017attention_transformer_TF,VIT} and the text stream of CLIP \cite{radford2021learning_CLIP} take question sentences and prompts as the input to extract text features in the target domain and general domain, respectively. 
2) Cross-domain Learning. After extracting the visual and language features separately in the target domain or in the general domain, these four sets of features are integrated through cross-domain learning. Four CLIP-guided visual-text encoders are designed for cross-domain cross-modal learning through the attention mechanism. Finally, the four sets of extracted features are fused through a multi-layer perceptron (MLP), and the answer decoder \cite{fan2019heterogeneous_HME_mem_att_memory_model_benchmark, le2022hierarchical_HCRN_relation_model_benchmark} is adopted to derive the correct answer. 

\subsection{Domain-specific Learning}
Given a video with a question to answer, the vision feature extraction, language feature extraction, and vision-language alignment are three critical steps for answering the question \cite{xu2017video_AMU_attention_model_benchmark,gao2018motion_CoMem_mem_memory_mode_benchmark,fan2019heterogeneous_HME_mem_att_memory_model_benchmark,jiang2020reasoning_HGA_graph_model_benchmark,jiang2020divide_QUEST_att_attention_model_benchmark,xiao2022video_HQGA_graph_model_benchmark,le2022hierarchical_HCRN_relation_model_benchmark}, while lack of general domain knowledge often makes the open-ended VideoQA very challenging \cite{lei2021less_VLP_Clipbert_benchmark,seo2021look_VLP_CoMVT_benchmark,yu2021learning_VLP_SSRea_benchmark}. To address this challenge, CLIP \cite{radford2021learning_CLIP} is integrated into the proposed method to incorporate the general domain knowledge. The proposed CCVQA model consists of two visual encoders and two language encoders. 

In the target domain, a 12-layer TimeSformer \cite{bertasius2021space_timesformer} is used to encode the spatial-temporal information embedded in each video frame. 
16 frames are randomly sampled from a video to preserve as much information as possible while keeping a low computational load. 
The TimeSformer partitions each frame into $N_{v}$ patches and generates patch tokens using a linear projection layer. After learnable positional embedding is added, the tokens are fed into the attention blocks to perform self-attention across the temporal and spatial dimensions. A temporal fusion layer aggregates the outputs from the attention blocks along the temporal dimension to obtain the sequenced features $\bm{H}_{v}=[\bm{v}_{cls}, \bm{v}_{1}, ..., \bm{v}_{N_{v}}]$, with $\bm{v}_{i} \in \mathbb{R}^{d}$. $d$ is the dimension of video features. $\bm{v}_{cls}$ is the classification token. 

In the target domain, a 6-layer BERT-base model \cite{devlin2018bert_BERT} is used to encode the language features $\bm{H}_{q}$. 
Given the input question of $N_{q}$ tokens, the BERT sequentially performs self-attention on the input and outputs an embedding sequence $\bm{H}_{q} = \left[\bm{q}_{cls},\bm{q}_{1},...,\bm{q}_{N_{q}}\right] \in \mathbb{R}^{N_{q} \times d} $, $\bm{q}_{i} \in \mathbb{R}^{d}$ and $\bm{q}_{cls}$ is the [CLS] token. $d$ is the dimension of question features, the same as that of video features. Learnable positional embeddings are added to the text tokens, similar to the video encoder. 

To incorporate general domain knowledge, a CLIP-guided visual encoder is designed. 
As it is time-consuming to encode the whole video sequence using CLIP, only key frames are selected according to the color histogram contrast\footnote{Code is available at \url{https://github.com/keplerlab/katna}} and encoded using the CLIP ViT-B/32 model \cite{VIT}. 
The CLIP is pre-trained on 400 million image-text pairs collected from the Internet, to map visual-text features into a joint embedding space, 
which can catch the similarity between encoded image features and linguistic features, and serve as additional language supervision to guide the subsequent text-image alignment.
The CLIP firstly transforms a key frame into $N_{k}$ patches using linear projections.
After adding the positional embedding, self-attentions are performed on the patches together with the [CLS] token to output the sequence features of the key frame $\bm{H}_{k}=\left[\bm{k}_{cls},\bm{k}_{1},...,\bm{k}_{N_{k}}\right]$, $\bm{k}_{i} \in \mathbb{R}^{d}$. 

The CLIP-guided text encoder is designed to 
extract the linguistic features using the general domain knowledge. The given question is used to generate the prompt, with the key information extracted from the question and the possible answer from the given answer space. Note that we are dealing with open-ended QA and hence there is no answer option but a very large space of $C$ possible answers. Similar to the CLIP-guided visual encoder, the CLIP language encoder tokenizes the prompt sentences, linearly projects them into the embedding space, and adds position embeddings with a layer normalization operation. The self-attention is then performed on the sentence embedding and produces a batch of feature vectors
$\bm{H}_{t} \in \mathbb{R}^{C \times N_{t} \times w}$,  where $N_{t}$ is the sequence length for each of the encoded prompt features 
$\left[\bm{t}_{1},...,\bm{t}_{N_{t}}\right]$, with 
$\bm{t}_{i} \in \mathbb{R}^{w}$, $w$ is the prompt feature dimensionality.




\subsection{Cross-domain Learning}
After deriving the visual features $\bm{H}_{v}$, $\bm{H}_{k}$ and linguistic features $\bm{H}_{q}$, $\bm{H}_{t}$ in a domain-specific way, four cross-domain cross-modal visual-text encoders are designed to capture the attentional information among these features. 
A 6-layer shared-weight transformer $\mathcal{T}$ is designed to model the attentional information between $\bm{H}_{q}$ and $\bm{H}_{v}$, and between $\bm{H}_{q}$ and $\bm{H}_{k}$. 
It directly takes the concatenated multi-modal features as the input, 
performs self-attention on paired visual features and linguistic features, and produces the class tokens with a MLP as formatted in Eqn. (\ref{eqn:1}) and (\ref{eqn:2}).
\begin{align}
    \bm{H}_{qv}&= \bm{W}_{qv}\mathcal{T}(\bm{H}_{q};\bm{H}_{v})+\bm{b}_{qv}, 
    \label{eqn:1} \\
    \bm{H}_{qk}&= \bm{W}_{qk}\mathcal{T}(\bm{H}_{q};\bm{H}_{k})+\bm{b}_{qk},
    \label{eqn:2}
\end{align}
where the class tokens $\bm{H}_{qv},\bm{H}_{qk} \in \mathbb{R}^{\bm{a}}$. $\bm{W}_{qv}$, $\bm{b}_{qv}$, $\bm{W}_{qk}$, and $\bm{b}_{qk}$ are trainable parameters. 

The size of CLIP encoded prompts $\bm{H}_{t} \in \mathbb{R}^{\bm{a} \times \bm{N}_{k} \times w}$ is much larger than that of the question features $\bm{H}_{q} \in \mathbb{R}^{\bm{N}_{q}\times d}$. The concatenation of $\bm{H}_{t}$ and $\bm{H}_{v}$ or $\bm{H}_{k}$ will result in large sequences and computing such sequences in transformer can take up to some GPU years \cite{radford2021learning_CLIP}. Therefore, to simplify the extraction of attentional information using self-attention via a transformer, we perform dot product for the interactions between the prompt text features $\bm{H}_{t}$ and the visual features $\bm{H}_{v}$ or $\bm{H}_{k}$. The [CLS] tokens of $\bm{H}_{t}$, $\bm{H}_{v}$ and $\bm{H}_{k}$ are linearly projected to vectors with the size of $\mathbb{R}^{\bm{a} \times w}$, $\mathbb{R}^{w}$ and $\mathbb{R}^{w}$, respectively. Then the two vectors are projected and multiplied to produce another two class tokens, 
\begin{align}
\bm{H}_{tv}=P(\bm{H}_{t}) \odot P(\bm{H}_{v}),\\ \bm{H}_{tk}=P(\bm{H}_{t}) \odot P(\bm{H}_{k}),
\end{align}
where $\bm{H}_{tv},\bm{H}_{tk} \in \mathbb{R}^{a}$, $\odot$ denotes the dot product, and $P(\cdot)$ denotes the projection. 

A weighted multi-head fusion is then utilized to integrate the four tokens $\bm{H}_{qv}$, $\bm{H}_{qk}$, $\bm{H}_{tv}$, and $\bm{H}_{qk}$. For each token, a linear layer is utilized for feature alignment so that each item on the token represents the confidence level for one class from the answer space. The final fused cross-domain cross-modal features $\bm{H}$ are derived as in Eqn. (\ref{eqn:5}), 
\begin{equation}    \bm{H}=\bm{W}_{1}\bm{H}_{qv}+\bm{W}_{2}\bm{H}_{qk}+\bm{W}_{3}\bm{H}_{tv}+\bm{W}_{4}\bm{H}_{tk}+\bm{b}
\label{eqn:5}
\end{equation}
where $\bm{W}_{1}$, $\bm{W}_{2}$, $\bm{W}_{3}$, $\bm{W}_{4}$, and $\bm{b}$ are trainable fusion weights.

\section{Experimental Results}



\subsection{Experimental Settings}

The proposed CCVQA is evaluated on two public datasets for open-ended VideoQA: MSVD-QA \cite{xu2017video_AMU_attention_model_benchmark} and MSRVTT-QA \cite{xu2017video_AMU_attention_model_benchmark}, which are generated using videos from the original datasets, MSVD \cite{chen2011collecting_MSVD} and MSRVTT \cite{xu2016msr_MSRVTT}, respectively, combined with auto-generated question answer annotation pairs. MSVD-QA contains 1.9K videos and 50k question-answer pairs in total and MSRVTT-QA has 10K videos and 243k question-answer pairs in total. The questions are categorized into five types: "what", "who", "when", "where", and "how" based on the starting words of the questions. The answers are all one-word nouns or verbs describing concrete objects or abstract concepts in the video. The standard dataset partition is followed \cite{xu2017video_AMU_attention_model_benchmark}. On the MSVD-QA dataset, $61\%$ of the data are used for training, $13\%$ for validation, and $26\%$ for testing. On the MSRVTT-QA dataset, the train-validation-test data split is $65\%$, $5\%$, and $30\%$. As for open-ended question answers, 
$2,423$ answer candidates are used for MSVD-QA and $1,500$ for MSRVTT-QA. 
The proposed method is compared with state-of-the-art models. AMU \cite{xu2017video_AMU_attention_model_benchmark}, Co-mem \cite{gao2018motion_CoMem_mem_memory_mode_benchmark}, HME \cite{fan2019heterogeneous_HME_mem_att_memory_model_benchmark}, HGA \cite{jiang2020reasoning_HGA_graph_model_benchmark}, SSML \cite{amrani2021noise_mm_learning_benchmark}, QUEST \cite{jiang2020divide_QUEST_att_attention_model_benchmark}, HCRN \cite{le2022hierarchical_HCRN_relation_model_benchmark}, 
STN \cite{yang2019question_benchmark}, DualVGR \cite{wang2021dualvgr_graph_model_benchmark}, and HQGA \cite{xiao2022video_HQGA_graph_model_benchmark}
 are selected as representative cross-modal learning methods. ClipBERT \cite{lei2021less_VLP_Clipbert_benchmark}, CoMVT \cite{seo2021look_VLP_CoMVT_benchmark}, SSRea \cite{yu2021learning_VLP_SSRea_benchmark}, 
 and VQA-T \cite{yang2021just_VLP_JustAsk_benchmark} are selected as typical pre-train model-enhanced methods. 

The CLIP ViT-B/32 architecture with pre-trained weights is obtained from the OpenAI's official release\footnote{CLIP model is available at \url{https://github.com/openai/CLIP}}. All video frames are resized to $224\times224$. 
On language feature encoding, the question is converted to the statement form at a size of $1,500$ sentences on MSRVTT-QA dataset and $2,423$ sentences on MSVD-QA dataset which are the same as the number of answer candidates. 
All experiments were performed on a single NVIDIA V100 GPU. 
The initial learning rate is set to $5\times10^{-5}$ and linearly decayed in the following epochs. The batch size is $24$ for the training processes on both datasets. The AdamW 
optimizer with a weight decay of $1\times10^{-3}$ is employed. The model converges within $15$ training epochs on MSVD-QA and $10$ epochs on MSRVTT-QA, consuming $120$ GPU hours and $400$ GPU hours respectively.


\subsection{Comparisons to State-of-the-art Models}
\vspace{-0.6em}
\begin{table}[htpb]
  \centering
  \resizebox{.8\columnwidth}{!}{
    \begin{tabular}{@{}lc  c}
    \toprule
    Methods & MSVD-QA & MSRVTT-QA \\
    \midrule
    AMU \cite{xu2017video_AMU_attention_model_benchmark} & 32.0 & 32.5       \\
    Co-mem \cite{gao2018motion_CoMem_mem_memory_mode_benchmark} & 31.7 & 32.0    \\
    HME \cite{fan2019heterogeneous_HME_mem_att_memory_model_benchmark} & 33.7 & 33.0       \\
    HGA \cite{jiang2020reasoning_HGA_graph_model_benchmark} & 34.7 & 35.5       \\
    SSML \cite{amrani2021noise_mm_learning_benchmark} & 35.13 & 35.06    \\
    QUEST \cite{jiang2020divide_QUEST_att_attention_model_benchmark} & 36.1 & 34.6     \\
    HCRN \cite{le2022hierarchical_HCRN_relation_model_benchmark} & 36.1 & 35.6     \\
    TSN \cite{yang2019question_benchmark} & 36.7 & 35.4    \\
    DualVGR \cite{wang2021dualvgr_graph_model_benchmark} & 39.03 & 35.52    \\
    HQGA \cite{xiao2022video_HQGA_graph_model_benchmark} & 41.2 & 38.6      \\
    \midrule
    ClipBERT \cite{lei2021less_VLP_Clipbert_benchmark} & - & 37.4     \\
    CoMVT \cite{seo2021look_VLP_CoMVT_benchmark} & 42.6 & 39.5     \\
    SSRea \cite{yu2021learning_VLP_SSRea_benchmark} & 45.5 & 41.6     \\
    \midrule                
    \textbf{Ours} & \textbf{46.6} & \textbf{42.4} \\
    \bottomrule
    \end{tabular}
    }
    \caption{Comparisons with state-of-the-art methods on MSRVTT-QA and MSVD-QA datasets in top-1 accuracy (\%).}
    \label{tab:1}
\end{table}

The comparisons with state-of-the-art methods are listed in Table \ref{tab:1}. The results confirm that our model outperforms all the compared methods by a considerable margin. The proposed method improves the top-1 accuracy from $45.5\%$ achieved by SSRea \cite{yu2021learning_VLP_SSRea_benchmark} to $46.6\%$ on the MSVD-QA dataset and from $41.6\%$ to $42.4\%$ on the MSRVTT-QA dataset, respectively. Compared with the VideoQA methods such as HQGA \cite{xiao2022video_HQGA_graph_model_benchmark} and DualVGR \cite{wang2021dualvgr_graph_model_benchmark} that do not use pre-trained models, the proposed model can significantly outperform them by nearly $5.4\%$ in accuracy on the MSVD-QA dataset and $3.8\%$ on the MSRVTT-QA dataset, which validates the effectiveness of using additional general domain knowledge. Compared to other pre-trained models such as SSRea \cite{yu2021learning_VLP_SSRea_benchmark} and CoMVT \cite{seo2021look_VLP_CoMVT_benchmark}, the proposed model further boosts the accuracy by around $1.1\%$ on the MSVD-QA dataset and $0.8\%$ on the MSRVTT-QA dataset, which demonstrates the effectiveness of our visual and linguistic features encoders and CLIP-guided visual-text attention mechanism. 


\subsection{Ablation Studies}
To show the performance gain brought about by each contribution, an ablation study is conducted. The proposed CCVQA is compared with two baselines: 1) \textbf{CCVQA w/o CLIP}, where the linguistic and visual clues from CLIP are not used, and 2) \textbf{CCVQA w/o Cross-domain Learning}, where only visual-text learning within the target/general domain is applied. As shown in Table \ref{tab:2}, CCVQA achieves a significant performance gain of 1.9\% and 1.1\% over \textbf{CCVQA w/o CLIP} on the MSVD-QA and MSVRTT-QA datasets, respectively, which demonstrates the effectiveness of our CLIP-guided design. CCVQA also achieves a performance gain of 0.7\% and 0.3\% through the proposed Cross-domain Learning on the MSVD-QA and MSVRTT-QA datasets, respectively.

\vspace{-0.3em}
\begin{table}[htbp]
    \centering
    \resizebox{1\columnwidth}{!}{
    \begin{tabular}{@{}lc  c}
    \toprule
        Method & MSVD-QA & MSVRTT-QA \\
    \midrule
        CCVQA w/o CLIP & 44.7 & 41.3 \\
        CCVQA w/o Cross-domain Learning & 45.9 & 42.1 \\
    \midrule    
    \textbf{CCVQA}  & \textbf{46.6} & \textbf{42.4} \\
    \bottomrule
    \end{tabular}
    }
    \caption{Ablation studies on the MSRVTT-QA and MSVD-QA datasets in terms of top-1 accuracy(\%).}
    \label{tab:2}
    \vspace{-1em}
\end{table}

\section{Conclusion}
To tackle the challenge of insufficient language supervision in VideoQA, a CLIP-guided cross-domain video-text encoder is proposed to transform CLIP's general domain knowledge into the application domain. Specifically, the pre-trained CLIP encodes key video frames and prompt texts using general domain knowledge. Together with video features extracted by the TimeSformer and question text features encoded by the BERT in the application domain, the attentional information across visual and text features are extracted using dot product and shared-weight transformer through cross-domain learning. Finally, the answer is decoded from the answer space. The proposed CCVQA is evaluated on two open-ended VideoQA datasets, MSVD-QA and MSRVTT-QA, which demonstrates a consistent and significant performance gain over the state-of-the-art models. 

\vfill\pagebreak

\vfill
\clearpage
\small
\balance
\bibliographystyle{bib/IEEEbib}
\bibliography{bib/strings,bib/refs}

\end{document}